\pdfoutput=1

\documentclass[11pt]{article}

 \usepackage[]{acl}

\usepackage{times}
\usepackage{latexsym}

\usepackage[T1]{fontenc}

\usepackage[utf8]{inputenc}

\usepackage{microtype}

\usepackage{multirow}
\usepackage{multicol}
\usepackage{graphicx}
\usepackage{subfigure}
\usepackage{xspace}
\usepackage{soul}
\usepackage{url}
\usepackage{enumerate}
\usepackage{enumitem}
\usepackage{booktabs}
\usepackage{amsmath}
\usepackage{color}
\usepackage{threeparttable}
\usepackage{bbding}
\usepackage{algorithm, algorithmic}

\usepackage{threeparttable}
\usepackage{dcolumn}
\newcolumntype{d}[1]{D{.}{.}{#1}}

\newcommand{\eat}[1]{}
\newcommand{\paratitle}[1]{\vspace{1ex}\noindent \textbf{#1}}

\let\oldhat\hat
\renewcommand{\vec}[1]{\mathbf{#1}}
\renewcommand{\hat}[1]{\oldhat{\mathbf{#1}}}
\renewcommand{\matrix}[1]{\mathbf{#1}}

\newcommand{\eg}{\emph{e.g.,}\xspace}

\newcommand{\ie}{\emph{i.e.,}\xspace}

\newcommand{\mynewsize}{\fontsize{9.5pt}{\baselineskip}{\selectfont}}

\title{BiSyn-GAT+: Bi-Syntax Aware Graph Attention Network for Aspect-based Sentiment Analysis}

\author{Shuo Liang\textsuperscript{ 1} , Wei Wei\textsuperscript{ 2, \Envelope}, Xian-Ling Mao\textsuperscript{ 3}, Fei Wang\textsuperscript{ 4}, Zhiyong He\textsuperscript{ 5}\\

  \textsuperscript{1,2} Cognitive Computing and Intelligent Information Processing (CCIIP) Laboratory, \\ 
  
  \mynewsize{School of Computer Science and Technology, Huazhong University of Science and Technology} \\
  
  \textsuperscript{3} School of Computer Science and Technology, Beijing Institute of Technology \\
  
  \textsuperscript{4} Institute of Computing Technology, Chinese Academy of Sciences \\
  \textsuperscript{5} Naval University of Engineering \\

  \textsuperscript{1} \texttt{shuoliang@hust.edu.cn}, \textsuperscript{2} \texttt{weiw@hust.edu.cn},
  \textsuperscript{3} \texttt{maoxl@bit.edu.cn},\\
 
  \textsuperscript{4} \texttt{wangfei@ict.ac.cn},
  \textsuperscript{5} \texttt{moonmon\_pub@outlook.com}
  \\
}

\begin{document}
\maketitle
\begin{abstract}
\let\thefootnote\relax\footnotetext{\Envelope ~ Corresponding Author}
Aspect-based sentiment analysis (ABSA) is a fine-grained sentiment analysis task that aims to align aspects and corresponding sentiments for aspect-specific sentiment polarity inference. It is challenging because a sentence may contain multiple aspects or complicated (\eg conditional, coordinating, or adversative) relations. Recently, exploiting dependency syntax information with graph neural networks has been the most popular trend. Despite its success, methods that heavily rely on the dependency tree pose challenges in accurately modeling the alignment of the aspects and their words indicative of sentiment, since the dependency tree may provide noisy signals of unrelated associations (\eg the ``\emph{conj}'' relation between ``\emph{great}'' and ``\emph{dreadful}'' in Figure \ref{Fig:dep_tree}).
In this paper, to alleviate this problem, we propose a \underline{\textbf{Bi}}-\underline{\textbf{Syn}}tax aware \underline{\textbf{G}}raph \underline{\textbf{At}}tention Network (\textbf{BiSyn-GAT+}).
Specifically, \textbf{BiSyn-GAT+} fully exploits the syntax information (\eg phrase segmentation and hierarchical structure) of the constituent tree of a sentence to model the sentiment-aware context of every single aspect (called \textbf{\emph{intra}}-context)
and the sentiment relations across aspects (called \textbf{\emph{inter}}-context) for learning.
Experiments on four benchmark datasets demonstrate that BiSyn-GAT+ outperforms the state-of-the-art methods consistently. 
\end{abstract}

\section{Introduction\label{sec:intro}}
\begin{figure}[!t]
	\centering
	\includegraphics[width=0.50\textwidth,height=0.13\textwidth]{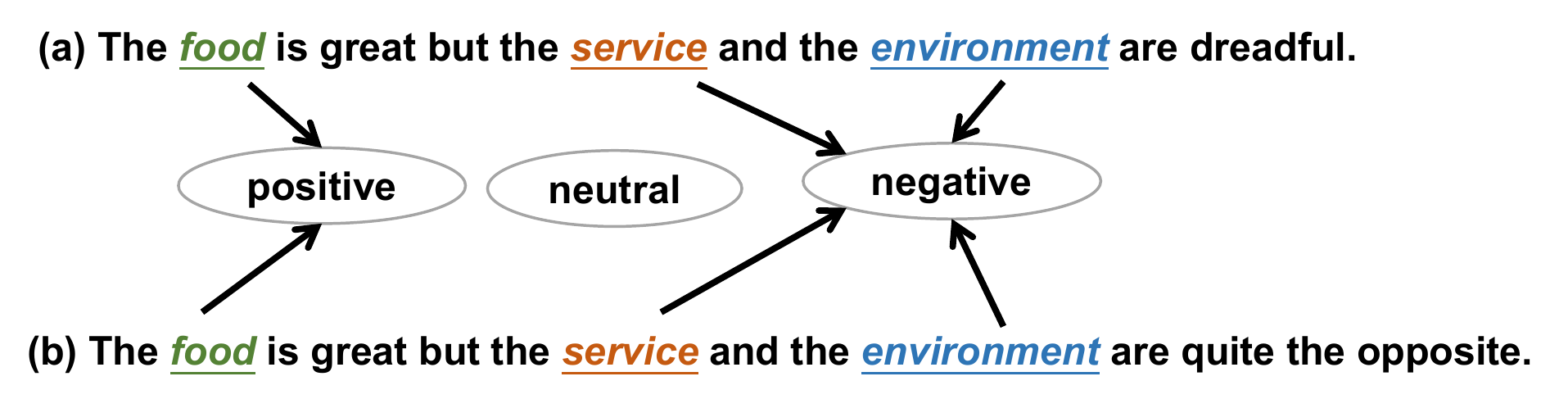}
	\vspace{-0.5cm}
	\caption{Examples of ABSA task. Each \underline{underlined} aspect is classified to corresponding sentiment polarity.} 
	\label{Fig:sentence examples}
	\vspace{-0.6cm}
\end{figure}

Aspect-based sentiment analysis (ABSA) aims to identify the sentiment polarity towards a given aspect in the sentence. Many previous works
\cite{Yang2018MultiEntityAS,Li2019ExploitingCT}
mainly focus on extracting sequence features via Recurrent Neural Networks (RNNs) or Convolution Neural Networks (CNNs) with attention mechanisms, which often assume that words closer to the target aspect are more likely to be related to its sentiment. However, the assumption might not be valid as exemplified in Figure \ref{Fig:sentence examples} (a), ``service'' is obviously closer to ``great'' rather than “dreadful”, and these methods may assign the irrelevant opinion word ``great'' to ``service'' mistakenly. 

\begin{figure}[!t]
	\centering
	\includegraphics[width=0.48\textwidth,height=0.27\textwidth]{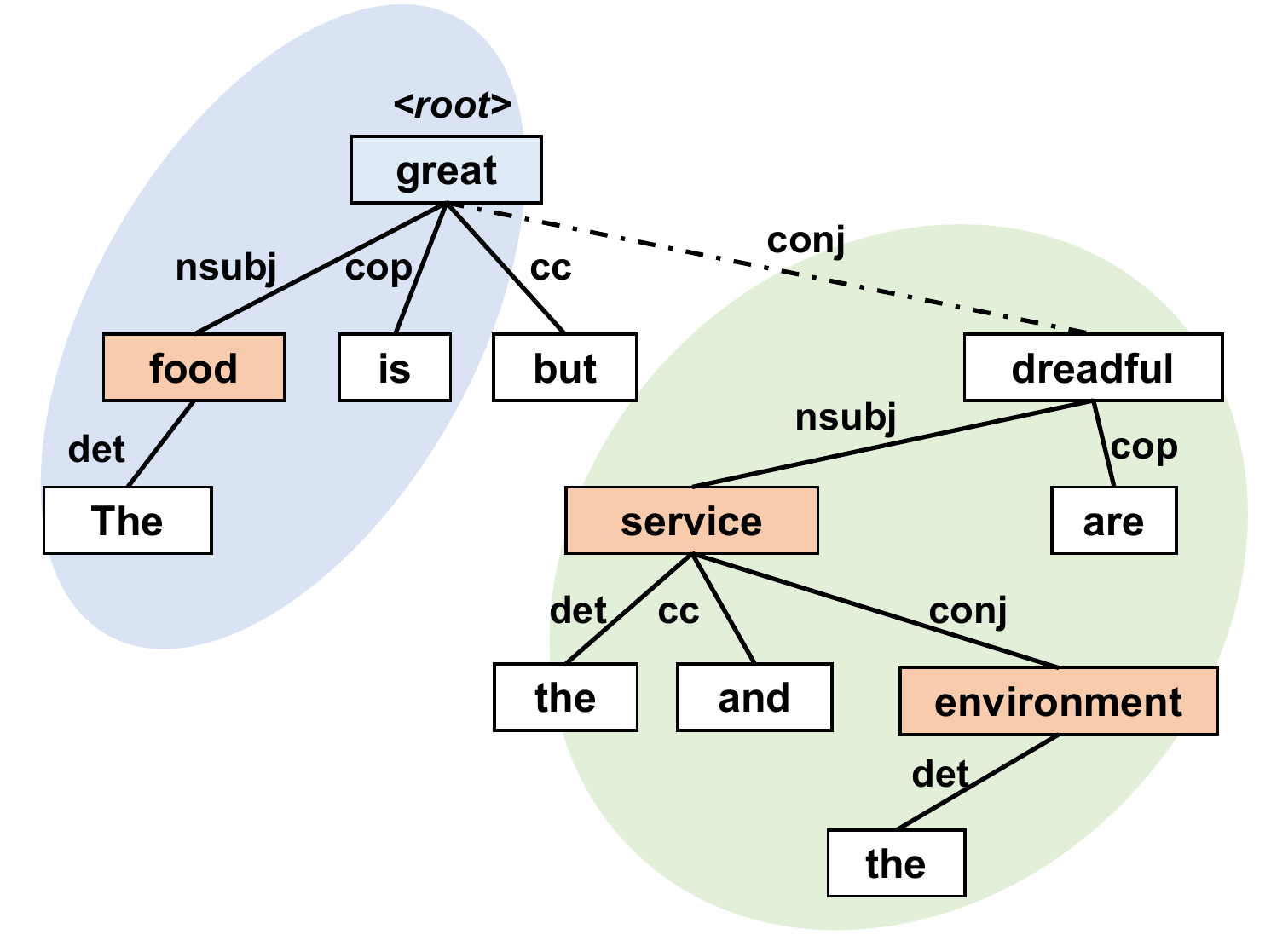}
	\vspace{-0.8cm}
	\caption{Dependency tree of ``The food is great but the service and the environment are dreadful''. Two separate ellipses encircle its two clauses. The ``conj'' edge between ``great'' and ``dreadful'' is a noise.
	} 
	\label{Fig:dep_tree}	
	\vspace{-0.8cm}
\end{figure}

To mitigate this problem, there already exists several efforts \cite{Wang2020RelationalGA,Chen2020InducingTL} dedicated to research on how to effectively leverage non-sequential information (\eg syntactic information like dependency tree) via Graph Neural Networks (GNNs). Generally, a dependency tree (\ie Dep.Tree), linking the aspect terms to the syntactically related words, stays valid in the long-distance dependency problem. 
However, the inherent nature of Dep.Tree structure may introduce noise like the unrelated relations across clauses, such as ``conj'' relation between ``great'' and ``dreadful" in Figure \ref{Fig:dep_tree}, which discourages capturing the sentiment-aware context of each aspect, \ie \emph{intra}-context.
Moreover, the Dep.Tree structure only reveals relations between words and, thereby, in most cases, is incapable of modeling complicated (\eg conditional, coordinating, or adversative) relations of sentences, therefore failing to capture sentiment relations between aspects, \ie \emph{inter}-context.

\begin{figure}
	\centering
	\includegraphics[width=0.5\textwidth,height=0.16\textwidth]{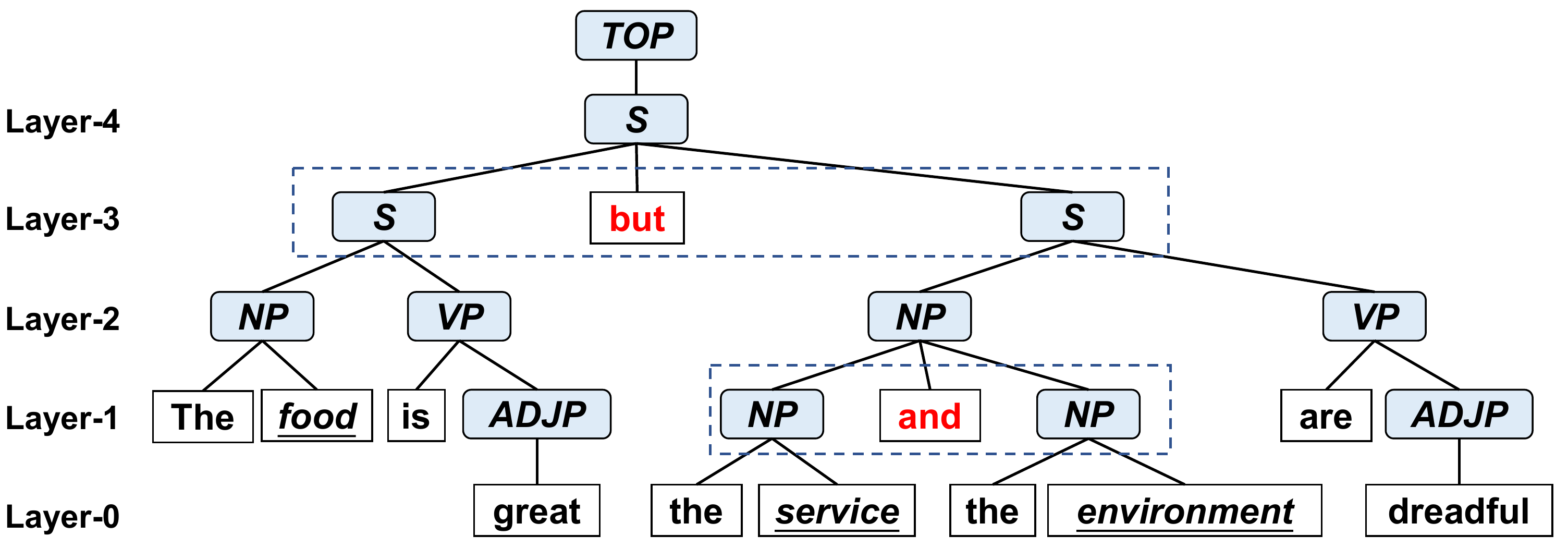}
	\vspace{-0.6cm}
	\caption{Constituent tree of the sentence ``The food is great but the service and the environment are dreadful''. Context words are in rectangles and parsed phrase types are in rounded rectangles.}
	\label{Fig:con_tree}
	\vspace{-0.5cm}
\end{figure}
Hence, in this paper, we consider fully exploiting the syntax information of the constituent tree to tackle the problem. 
Typically, a constituent tree (\ie Con.Tree) often contains precise and discriminative phrase segmentation and hierarchical composition structure, which are helpful for correctly aligning the aspects and their corresponding words indicative of sentiment.
The former can naturally divide a complicated sentence into multiple clauses, and the latter can discriminate different relations among aspects to infer the sentiment relations of different aspects.
We illustrate this with an example in Figure \ref{Fig:con_tree}: (1) Clause ``The food is great'' and the clause ``the service and environment are dreadful'' are segmented by the phrase segmentation term ``but''; (2) In Layer-1, the term ``and'' indicates the coordinating relation of ``service'' and ``environment'', while the term ``but'' in Layer-3 reflects the adversative relation towards ``food'' and ``service'' (or ``environment'').

Thus, to better align aspect terms and corresponding sentiments, we propose a new framework, \underline{\textbf{Bi}}-\underline{\textbf{Syn}}tax aware \underline{\textbf{G}}raph \underline{\textbf{At}}tention Network (\textbf{BiSyn-GAT+}), to effectively leverage the syntax information of constituent tree by modeling \emph{intra}-context and \emph{inter}-context information. In particular, BiSyn-GAT+ employs: 1) a syntax graph embedding to encode the \emph{intra}-context of each aspect based on the fusion syntax information within the same clause in a bottom-up way, which combines the phrase-level syntax information of its constituent tree and the clause-level syntax information of its dependency tree. 2) an aspect-context graph consisting of phrase segmentation terms and all aspects to model the \emph{inter}-context of each aspect. Specifically, it aggregates the sentiment information of other aspects according to the influence between the current aspect and its neighbor aspects, which is calculated based on aspect representations learned from bi-directional relations over the aspect context graph, respectively.

Our main contributions are as follows:

(1) To the best of our knowledge, this is the first work to exploit syntax information of constituent tree (\eg phrase segmentation and hierarchical structure) with GNNs for ABSA. Moreover, it shows superiority in the alignments between aspects and corresponding words indicative of sentiment.

(2) We propose a framework, \underline{\textbf{Bi}}-\underline{\textbf{Syn}}tax aware \underline{\textbf{G}}raph \underline{\textbf{At}}tention Network (\textbf{BiSyn-GAT+}), to fully leverage syntax information of constituent tree (or, and dependency tree) by modeling the sentiment-aware context of each single aspect and the sentiment relations across aspects.

(3) Extensive experiments on four datasets show that our proposed model achieves state-of-the-art performances.
\section{Related Work}
Sentiment analysis is an important task in the field of natural language processing~\cite{zhang2018deep, yang2020cm} and can be applied in downstream tasks, like emotional chatbot~\cite{wei2019emotion, li-etal-2020-empdg, 10.1145/3423168, wei2021target}, recommendation system~\cite{zhao2022multi, wang2020global}, QA system ~\cite{Wei2011IntegratingCQ,Qiu2021ReinforcedHB}. Here we focus on a fine-grained sentiment analysis task — ABSA.
Recently, deep learning methods have been widely adopted for ABSA task. These works can be divided into two main categories: methods without syntax information (\ie Syntax-free methods) and methods with syntax information (\ie Syntax-based methods).

\paratitle{Syntax-free methods}:
Neural networks with attention mechanisms \cite{wang-etal-2016-attention,chen-etal-2017-recurrent,Song2019AttentionalEN} have been widely used.
\citet{chen-etal-2017-recurrent} adopts a multiple-attention mechanism to capture sentiment features.
\citet{Song2019AttentionalEN} uses an attentional encoder network (AEN) to excavate rich semantic information from word embeddings.

\paratitle{Syntax-based methods}:
Recently, utilizing dependency information with GNNs has become an effective way for ABSA.
\citet{zhang-etal-2019-aspect} uses graph convolutional networks (GCN) to learn node representations from Dep.Tree. 
\citet{tang-etal-2020-dependency} proposes a dependency graph enhanced dual-transformer network (DGEDT) by jointly considering representations from Transformers and corresponding dependency graph.  
\citet{Wang2020RelationalGA} constructs aspect-oriented dependency trees and proposes R-GAT, extending the graph attention network to encode graphs with labeled edges.
\citet{li-etal-2021-dual-graph} proposes a dual graph convolutional networks (DualGCN) model,  simultaneously considering syntax structures and semantic correlations.
All above works use syntax information of Dep.Tree, which may introduce noise, as we said before. Thus, we exploit syntax information of Con.Tree with GNNs. Precisely, we follow the Con.Tree to aggregate information from words within the same phrases in a bottom-up way and capture \emph{intra}-context information.

Moreover, some works resort to modeling aspect-aspect relations. Some \cite{hazarika2018modeling,majumder-etal-2018-iarm} adopt aspect representations to model relations by RNNs or memory networks, without utilizing context information. And some \cite{fan-etal-2018-multi,hu-etal-2019-constrained} propose alignment loss or orthogonal attention regulation to constrain aspect-level interactions, which fail when aspects have no explicit opinion expressions or multiple aspects share same opinion words. Recently, there are some works utilizing GNNs to model aspect relations.
\citet{liang-etal-2020-jointly} constructs an inter-aspect graph based on relative dependencies between aspects. 
\citet{ZHAO2020105443} constructs a sentiment graph, where each node represents an aspect, and each edge represents the sentiment dependency relation.
However, these works fail to explicitly use phrase segmentation information, such as conjunction words. Thus, we propose an aspect-context graph consisting of all aspects and phrase segmentation terms to model \emph{inter}-context information.

\paratitle{GNNs with constituent tree}:
To our knowledge, we are the first work to utilize the constituent tree for ABSA task. But in aspect-category sentiment analysis task, which predicts sentiment polarity towards a given predefined category in the text, \citet{Li2020SentenceCA} proposes a Sentence Constituent-Aware Network (SCAN) that generates representations of the nodes in Con.Tree. Unlike SCAN, we view parsed phrases as different spans of the input text instead of individual nodes. So we don't introduce any inner nodes of Con.Tree ({\eg} ``NP'',``VP'' of Figure~\ref{Fig:con_tree}) into the representation space, decreasing the computational overhead.
\begin{figure*}[!t]
	\centering
	\includegraphics[width=.9\textwidth,height=0.5\textwidth]{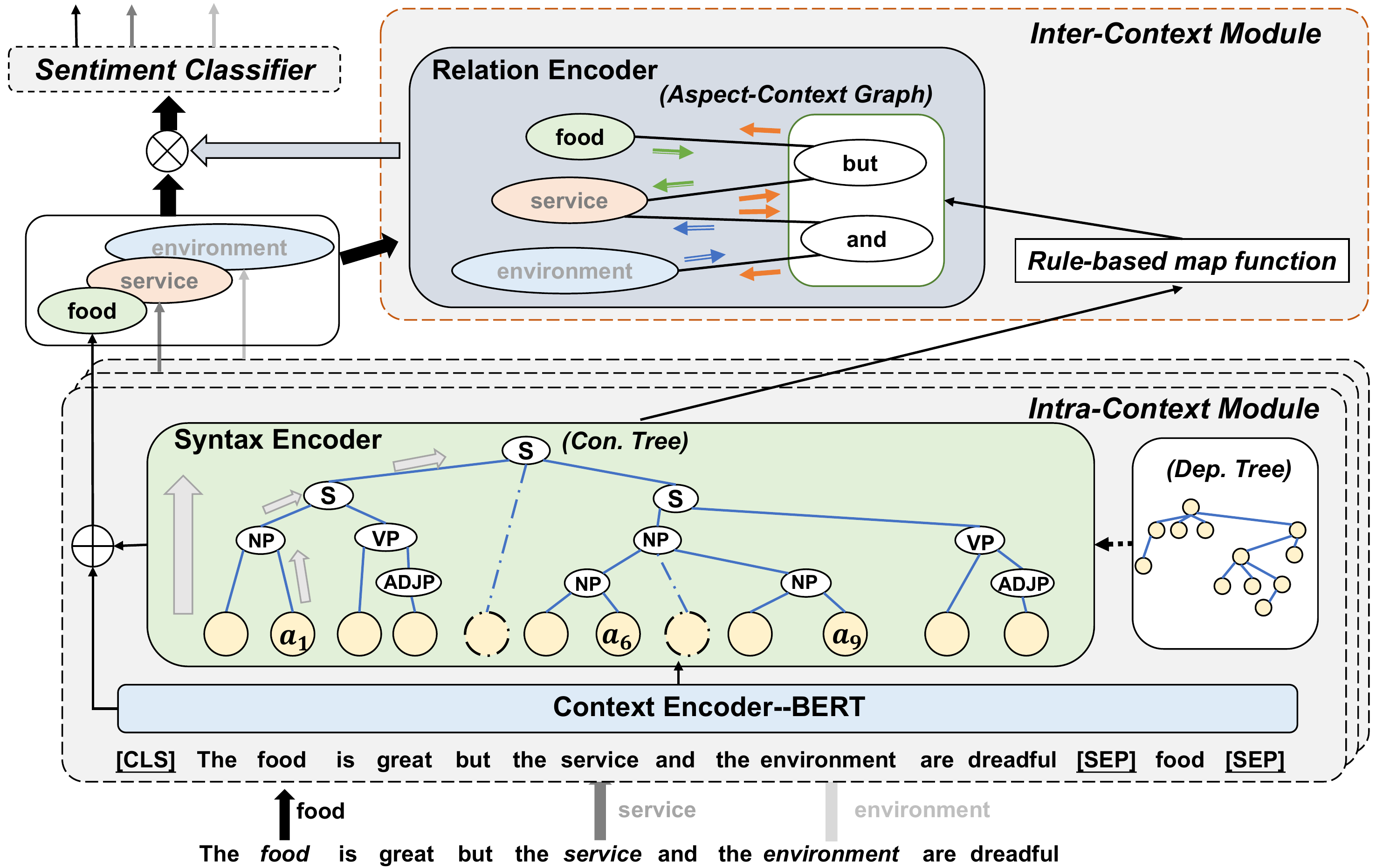}
	 \vspace{-0.2cm}
	\caption{
		Overall architecture. It takes the sentence and all aspects as input and outputs sentiment predictions for all aspects.
		It has three components: 1) the \textbf{\emph{intra}-context module} contains two encoders: a \underline{context encoder} that outputs contextual word representations and a \underline{syntax encoder} that utilizes syntax information of the parsed constituent tree (or, and dependency tree). Output representations from two encoders are fused to generate aspect-specific representations; 2) the \textbf{\emph{inter}-context module} includes a \underline{relation encoder} applied to the constructed aspect-context graph to obtain relation-enhanced representations. The aspect-context graph includes all aspects and phrase segmentation terms obtained from a designed rule-based map function applied to the constituent tree. 3) the \textbf{sentiment classifier} takes the outputs from two modules to make predictions.} 
	\label{Fig:Framework}
	 \vspace{-0.4cm}
\end{figure*}
\section{Methodology}

\subsection{Overview}
\paratitle{Problem Statement}.
Let $\vec{s}=\left\{w_i\right\}_{n}$ and $\matrix{A} = \left\{a_j \right\}_{m}$ be a sentence and a predefined aspect set,
where $n$ and $m$ are the number of words in $\vec{s}$
and
the number of aspects in $\matrix{A}$, respectively.
For each $\vec{s}$, $\matrix{A_s} = \left\{a_i| a_i\in{\matrix{A}}, a_i \in \vec{s} \right\}$
denotes the aspects contained in $\vec{s}$. We treat each multiple-word aspect as a single word for simplicity, so $a_i$ also means the $i$-th word of $\vec{s}$.
The goal of ABSA is to predict the sentiment polarity
$y_i\in \left\{\text{positive, negative, neural}\right\}$ for each aspect $a_i\in \matrix{A_s}$.
 
\paratitle{Architecture}.
As shown in Figure \ref{Fig:Framework}, our proposed architecture takes the sentence and all aspects that appear in the text as the input, and outputs the sentiment predictions of the aspects. It contains three components:
1) the \textbf{\emph{intra}-context module} encodes the input $\left\{w_i\right\}$ to obtain aspect-specific representations of the target aspects, which contains two encoders: a \underline{context encoder} that outputs contextual word representations and a \underline{syntax encoder} that utilizes syntax information of the parsed constituent tree (or, and dependency tree).
2)  the \textbf{\emph{inter}-context module} includes a \underline{relation encoder} applied to the constructed aspect-context graph to output relation-enhanced representations. The aspect-context graph composes all aspects of the given sentence and phrase segmentation terms obtained from a designed rule-based map function applied to the constituent tree.
3) the \textbf{sentiment classifier} takes output representations of the above two modules to make predictions.

\subsection{\emph{Intra}-Context Module}
In this part, we utilize a context encoder and a syntax encoder to model the sentiment-aware context of every single aspect and generate aspect-specific representation for each aspect. Note that for multi-aspect sentences, we use this module multiple times, as each time deals with one aspect.
\subsubsection{Context Encoder}
We use BERT~\cite{Devlin2019BERTPO} to generate contextual word representations. Given target aspect $a_t$, we follow \textbf{BERT-SPC}~\cite{Song2019AttentionalEN} to construct a BERT-based sequence:
\begin{equation}
BERT\_seq_t= \left[\text{CLS}\right] + \left\{w_i\right\} + \left[\text{SEP}\right] + a_t + \left[\text{SEP}\right],
\end{equation}
Then, the output representation is obtained by,
\begin{equation}
	\begin{gathered}
		h^{t}=\left\{h_{0}^{t}, h_{1}^{t}, \ldots, h_{n^{\prime}}^{t}, \ldots, h^{t}_{n^{\prime}+2+m_{t}^{\prime}} \right\}
	\end{gathered}
\end{equation}
where $n^{\prime}$ and $m^{\prime}$ are lengths of input text and target aspect $a_t$ after BERT tokenizer separately, $h_0^{t}$ is ``BERT pooling'' vector representing the BERT sequence, $h_i^t$ is the contextual representation of each token. Note that $w_i$ may be split into multiple sub-words by BERT tokenizer. So we calculate the contextual representation of $w_i$ as follows, 
\begin{equation}
	\begin{gathered}
		\hat{h}^{t}_{i}= \frac{1}{|BertT(w_i)|}\sum_{k \in BertT(w_i) }h_k^{t} ,
	\end{gathered}
\end{equation}
where $BertT(w_i)$ returns an index set of $w_i$'s sub-words in BERT sequence, and $|\ |$ returns its length.
\subsubsection{Syntax Encoder}
The above representations only consider semantic information, so we propose a syntax encoder to utilize rich syntax information. Our syntax encoder is stacked by several designed \underline{H}ierarchical \underline{G}raph \underline{AT}tention (HGAT) blocks, and each block consists of multiple graph attention (\ie GAT) layers that encode syntax information hierarchically under the guidance of the constituent tree (or, and the dependency tree). The key point is the construction of corresponding graphs.

\begin{figure}[!t]
	\centering
	\includegraphics[width=0.5\textwidth,height=0.5\textwidth]{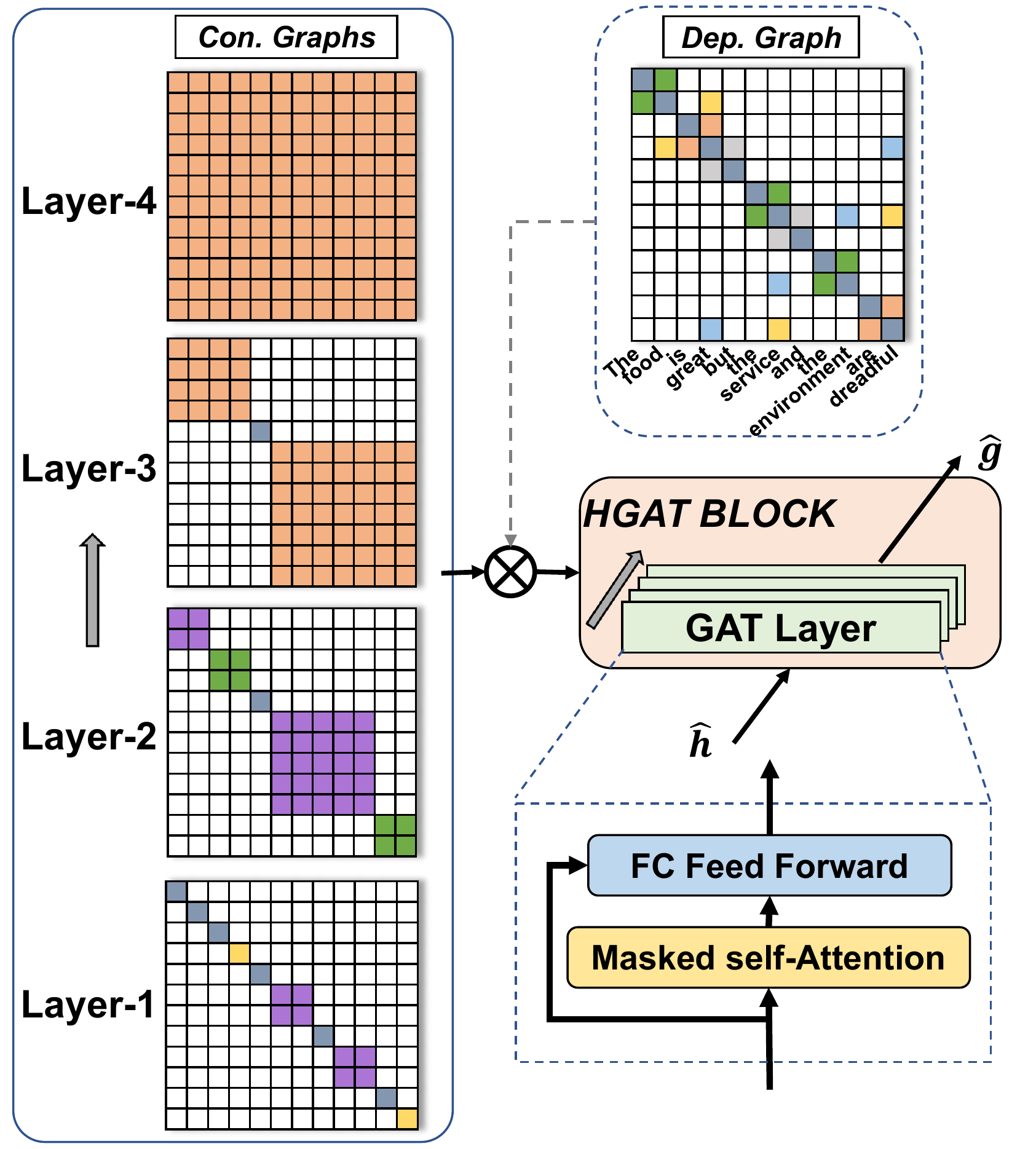}
	 \vspace{-0.5cm}
	\caption{HGAT Block. It is stacked by several GAT layers, and each GAT layer is applied to the graph obtained from one layer of the constituent tree (or, and the dependency tree).}
	\label{Fig:HGAT Block}
	 \vspace{-0.5cm}
\end{figure}

\paratitle{Graph construction}.
As Figure \ref{Fig:Framework} shows, we follow the syntax structure of Con.Tree in a bottom-up way. Each layer $l$ of Con.Tree consists of several phrases $\left\{ph_u^l\right\}$ that compose the input text, and each phrase represents an individual semantic unit. \eg $\left\{ph^3\right\}$ in Figure ~\ref{Fig:con_tree} is \{The food is great, but, the service and the environment are dreadful\}.
We construct corresponding graphs based on those phrases. \ie
For layer $l$ that consists of phrases $\left\{ph_{u}^{l}\right\}$, we construct the adjacent matrix $\matrix{CA}$ that shows word connections:
\begin{equation}
	\matrix{CA}_{i, j}^{l}= \begin{cases}
		1 & \small{ \text{if ~} w_i,w_j \text{ in same phrase of} \left\{ph_{u}^{l}\right\}} \\ 
		0 & \small{ \text{otherwise }}
	\end{cases},
\end{equation}
which is exemplified as \emph{Con.Graphs} in Figure \ref{Fig:HGAT Block}.

\paratitle{HGAT block}. A HGAT block aims to encode syntax information into word representations hierarchically. As Figure \ref{Fig:HGAT Block} shows, a HGAT block is stacked by several GAT layers that utilize a masked self-attention mechanism to aggregate information from neighbors and a fully connected feed forward network to map representations to the same semantic space. Attention mechanism can handle the diversity of neighbors with higher weights assigned to more related words. It can be formulated as follows,
\begin{equation}
	\hat{\vec{g}}_{i}^{t,l} = FC(\vec{g}_{i}^{t,l} + \hat{\vec{g}}_i^{t,l-1}),
\end{equation}
\begin{equation}
	\vec{g}_{i}^{t,l}=\|_{z=1}^{Z} \sigma\left(\sum_{j \in \mathcal{N}^{t,l}(i)} \alpha_{i j}^{l z} \matrix{W}_{g}^{l z} \hat{\vec{g}}_{j}^{t,l-1}\right),
\end{equation}
\begin{equation}
	\alpha_{i j}^{l z}=\frac{\exp \left(f\left(\hat{\vec{g}}_{i}^{t,l-1}, \hat{\vec{g}}_{j}^{t,l-1}\right)\right)}{\sum_{j^{\prime} \in \mathcal{N}^{l}(i)} \exp \left(f\left(\hat{\vec{g}}_{i}^{t,l-1}, \hat{\vec{g}}_{j^{\prime}}^{t,l-1}\right)\right)},
	\label{score_function}
\end{equation}
where $\mathcal{N}^{l}(i)$ is the set of neighbors of $w_i$ in layer $l$, $\hat{\vec{g}}_i^{t,l}$ is the final representation of $w_i$ in layer $l$, $FC$ is fully connected feed forward network. $\vec{g}_i^{t,l}$ is the representation of $w_i$ after masked self-attention mechanism.  $||$ denotes vector concatenation. $Z$ is the number of attention heads, $\sigma$ is activation function. $W_g^{lz}$ is trainable parameter of the $z$th head of layer $l$. $f$ is a score function that measures the correlation of two words. Stacked HGAT block takes the output of previous one as the input, and the input of the first HGAT block is $\hat{h}^{t}$. The output of syntax encoder is defined as $\hat{g}^{t}$ for simplicity.

\paratitle{With dependency information.} We also explore the fusion of two syntax information. Following previous works, we consider the Dep.Tree as an undirected graph and construct adjacent matrix $\matrix{DA}$, which is formulated as follows,
\begin{equation}
	\matrix{DA}_{i, j} = \begin{cases}
		1 & \text { if } w_i,w_j \text{ link directly in Dep.Tree} \\ 
		0 & \text { otherwise }
	\end{cases}
\end{equation}

We consider three operations: \textbf{position-wise dot}, \textbf{position-wise add}, and \textbf{conditional position-wise add}. Each corresponding adjacent matrix $\matrix{FA}$ is shown as follows,

\paratitle{A. position-wise dot}. For each layer of Con.Tree, this operation only considers neighbors of the Dep.Tree that are also in the same phrase.
\begin{equation}
	\matrix{FA} = \matrix{CA} \cdot \matrix{DA} 
\end{equation}

\paratitle{B. position-wise add}. For each layer of Con.Tree, this operation considers words in the same phrases and neighbors of the Dep.Tree. Some edges of Dep.Tree can shorten paths between aspect words and relevant opinion words, {\eg} ``food'' and ``great'' in Figure ~\ref{Fig:con_tree}. 
\begin{equation}
\matrix{FA} = \matrix{CA} + \matrix{DA} 
\end{equation}

\paratitle{C. conditional position-wise add}. This operation considers phrase-level syntax information of Con.Tree and clause-level syntax information of Dep.Tree. Specifically, it first deletes all dependency edges that are across clauses (\eg the edge between ``great'' and ``dreadful'' in Figure \ref{Fig:dep_tree}) and then conducts \textbf{position-wise add} operation with the remaining dependency edges. 
\begin{equation}
	\matrix{FA} = \matrix{CA} \oplus \matrix{DA} 
\end{equation}

Thus, the output of the \emph{intra}-context module contains both contextual information and syntax information, which is formulated as follows,
\begin{equation}
	\vec{v}_t^{as} = \left[\hat{\vec{h}}_t^{t} + \hat{\vec{g}}_t^{t} ; h_0^{t}\right]
\end{equation}

\begin{figure}[b]
	\centering
	\includegraphics[width=0.32\textwidth,height=0.28\textwidth]{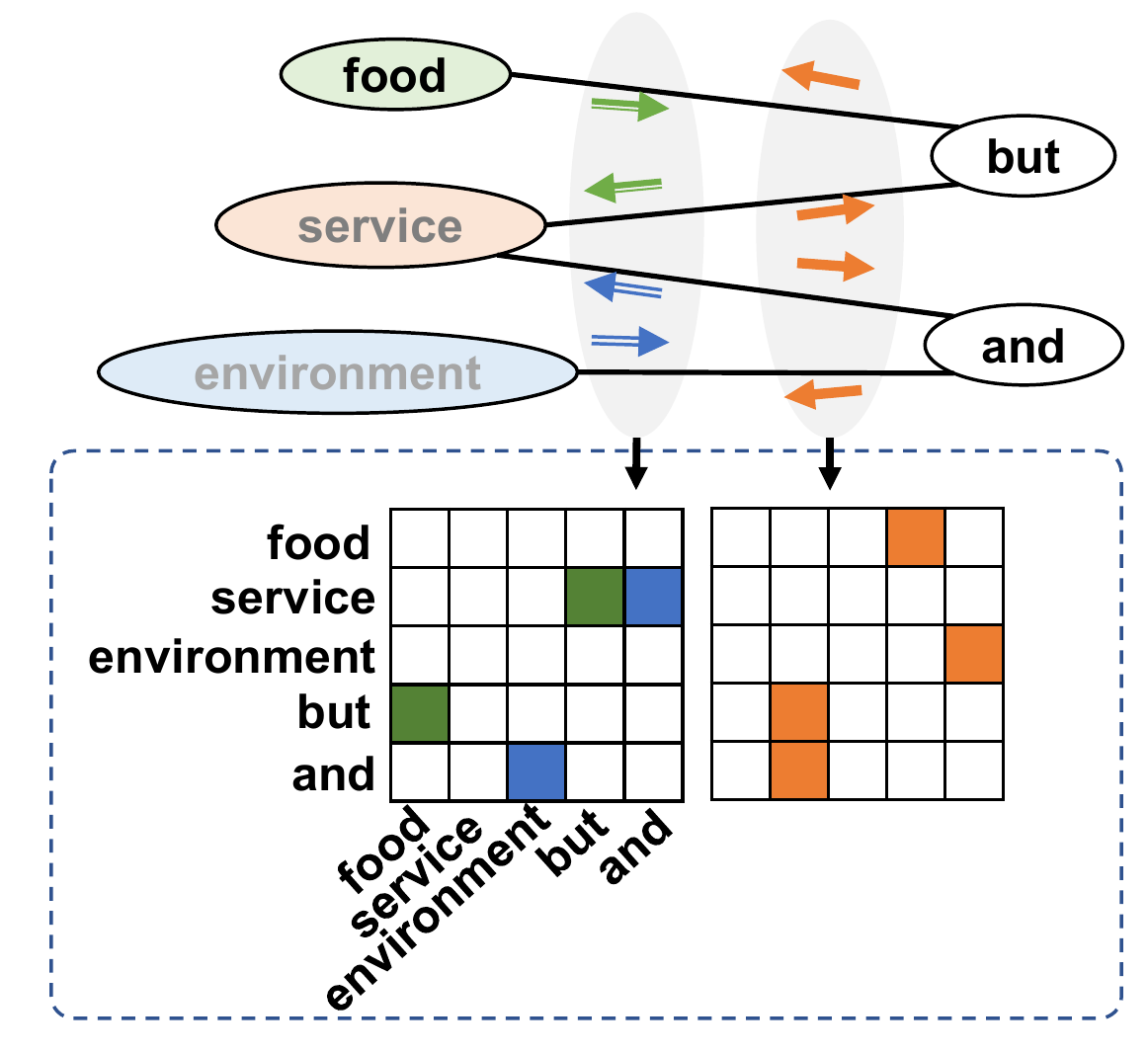}
	\vspace{-0.2cm}
	\caption{Example of an aspect-context graph and corresponding two adjacent matrices for distinguishing the bi-directional relations.}
	\label{Fig:aspect_context_graph}
	\vspace{-0.6cm}
\end{figure}

\subsection{\emph{Inter}-Context Module}

The \emph{intra}-context module ignores the mutual influence of aspects. Thus, in \emph{inter}-context module, we construct an aspect-context graph to model the relations across aspects. This module only works for multi-aspect sentences, with aspect-specific representations of all aspects from \emph{intra}-context module as input and outputs relation-enhanced representation of each aspect. 

\paratitle{Phrase segmentation.} Aspect relations can be revealed by some phrase segmentation terms, like conjunction words. Thus, we design a rule-based map function \textbf{$PS$} that returns phrase segmentation terms of two aspects: Given two aspects, it first finds their lowest common ancestor (LCA) in the Con.Tree, which contains information of two aspects and has the least irrelevant context. We call branches from LCA that between sub-trees which two aspects are separately in as ``inner branches''. \textbf{$PS$} returns all text words in the inner branches if they exist; else, it returns words between two aspects of the input text. It is formulated as follows,
\begin{equation}
	PS(a_i,a_j) =  \begin{cases}		
		\left\{w_k\right\}, & \text{if } |Br(a_i,a_j)|=0 \\
		Br(a_i,a_j),  &  otherwise
	\end{cases},
\end{equation}
where $i < k < j$ and $Br(a_i,a_j)$ returns text words in the inner branches of $a_i$ and $a_j$. {\eg} in Figure \ref{Fig:con_tree}, given aspects food and service, the LCA node is \emph{S} of Layer-4 that has three branches, with food in the first and service in the third. So ``but'' in the second branch (inner branch) is the phrase segmentation term that reflects sentiment relation of two aspects.

\paratitle{Aspect-context graph construction.} We notice that the influence range of one aspect should be continuous, and the mutual influence of aspects attenuates with distance. Considering all aspect pairs introduces noise caused by long distance and increases computational overhead. So we only model relations across neighbor aspects. 
After extracting phrase segmentation terms of neighbor aspects by \textbf{$PS$} function, we construct an aspect-context graph by linking aspects with corresponding phrase segmentation terms to help infer relations. To distinguish the bi-directional relations over the aspect-context graph, we build two corresponding adjacent matrices. The first handles influence from aspects in odd-index among all aspects of the sentence, to neighbor even-index aspects, the second handles the opposite. An example is shown in Figure \ref{Fig:aspect_context_graph}. Then, taking $\left\{v^{as}_t, t\in \matrix{A_s}\right\}$ and corresponding phrase segmentation terms representations encoded by BERT as the input, the above HGAT blocks are applied as the relation encoder to obtain relation-enhanced representation $v^{aa}_t$ for each aspect $a_t$.

\subsection{Training}
The outputs of the \textbf{\emph{intra}-context module} and \textbf{\emph{inter}-context module} are combined to form the final representations, which are  later fed to a fully connected layer (\ie sentiment classifier) with a softmax activation function, generating the probabilities over the three sentiment polarities:
\begin{gather}
    \vec{o_t} = \vec{v}_t^{as} + \vec{v}_t^{aa},\label{eq:o_t}\\
    \vec{p(t)} = softmax(\matrix{W_p}\vec{o_t}+\vec{b_p}),
\end{gather}
where $\matrix{W_p}$, $\vec{b_p}$ are parameters of the classifier\footnote{In Eq\ref{eq:o_t}, $\vec{v}_t^{aa}$ is set to zero in single-aspect sentence.}.

The loss is defined as the cross-entropy loss between golden polarity labels and predicted polarity distributions of all (sentence, aspect) pairs:
\begin{equation}
L (\theta)^{Sentiment} =-\sum_{s} \sum_{a_t \in A_s} loss(\vec{p(t)},y(t)),
\end{equation}
where $a_t$ is the aspect and also the $t$-th word in $s$, $loss$ is the standard cross-entropy loss, $\theta$ represents model parameters.

\begin{table}[t]\footnotesize
	\centering
	\renewcommand\arraystretch{0.95}
	\setlength{\tabcolsep}{0.92pt}
	\begin{tabular}{cc|ccc|ccc}
		\hline
		\multicolumn{2}{c|}{\textbf{Dataset}} & \multicolumn{3}{c|}{\textbf{Sentence-Level}} & 
		\multicolumn{3}{c}{\textbf{Aspect-Level}} \\
		\cline{3-8}
		~ & ~ & {\textbf{Multi-Asp.}} & {\textbf{Single-Asp.}} & {\textbf{All}} & {\textbf{Pos.}} & {\textbf{Neg.}} & {\textbf{Neu.}} \\
		\hline
		{Rest-} & Train & 971 & 1009 & 1980 & 2164 & 807 & 637 \\
		{aurant} & Test  & 315 & 284 & 599 & 727 & 196 & 196 \\
		\hline
		\multirow{2}*{Laptop} & Train & 538 & 916 & 1454 & 937 & 851 & 455  \\
		~ & Test  & 150 & 259 & 409 & 337 & 128 & 167\\
		\hline
		\multirow{3}*{MAMS} & Train  & 4297 & 0 & 4297 & 3380 & 2764 & 5042 \\
		~ & valid & 500 & 0 & 500 & 403 & 325 & 604 \\
		~ & Test & 500 & 0 & 500 & 400 & 329 & 607 \\
		\hline
		\multirow{2}*{Twitter} & Train  & 0 & 6051 & 6051 & 1507 & 1528 & 3016 \\
		~ & Test & 0 & 677 & 677 & 172 & 169 & 336 \\
		\hline
	\end{tabular}
	\vspace{-0.2cm}
	\caption{\label{Table:dataset}
		Statistics of datasets. Multi-Asp., Single-Asp. indicate the number of sentences with multiple or single aspect; Pos., Neg., and Neu. show the number of aspects towards positive, negative and neutral label.
	}
	\vspace{-0.4cm}
\end{table}

\begin{table*}[!t]\footnotesize
	\centering
	\begin{tabular}{cc|cc|cc|cc|cc}
		\hline
		\multirow{3}*{\textbf{Category}} &  \multirow{3}*{\textbf{Model}} &  \multicolumn{8}{c}{\textbf{Dataset}}  \\
		\cline{3-10}
		~ & ~ & \multicolumn{2}{c|}{\textbf{Restaurant}} &  \multicolumn{2}{c|}{\textbf{Laptop}}  &  \multicolumn{2}{c|}{\textbf{MAMS}} &  \multicolumn{2}{c}{\textbf{Twitter}}\\
		~ & ~ & Acc.(\%)&F1.(\%) & Acc.(\%) &F1.(\%)  & Acc.(\%) &F1.(\%) & Acc.(\%) &F1.(\%) \\
		\hline 
		w/o Syn. & BERT-SPC & 84.46 & 76.98 & 78.99 & 75.03 & 82.82 & 81.90 & 73.55 & 72.14 \\
		& AEN-BERT & 83.12 & 73.76 & 79.93 & 76.31 & - & - & 74.71 & 73.13 \\
		\hline
		w/ Syn. & R-GAT & 86.60 & 81.35 & 78.21 & 74.07 & - & - & 76.15 & 74.88 \\
		& RGAT+ & 86.68 & 80.92 & 80.94 & 78.20 & 84.52 & 83.74 & 76.28 & 75.25 \\ 
		& DGEDT & 86.30 & 80.00 & 79.80 & 75.60 & - & - & \underline{77.90} & 75.40 \\
		& DualGCN & 87.13 & 81.16 & 81.80 & 78.10 & - & - & 77.40 & \underline{76.02} \\
		\cline{2-10}
		& SDGCN & 83.57 & 76.47 & 81.35 & 78.34 & - & - & - & - \\
		& InterGCN & 87.12 & 81.02 & \underline{82.87} & \underline{79.32} & - & - & - & - \\
		\hline
		Ours & BiSyn-GAT &  \underline{87.49} & \underline{81.63} & 82.44 & 79.15 & \underline{84.90} & \underline{84.43} & \textbf{77.99} & \textbf{76.80} \\
		& BiSyn-GAT+ & \textbf{87.94} & \textbf{82.43} & \textbf{82.91} & \textbf{79.38}  & \textbf{85.85} & \textbf{85.49} &  &  \\
		\hline
	\end{tabular}
	\vspace{-0.25cm}
	\caption{\label{Table:exp-result} Performance comparison of  models on four datasets. The best are in \textbf{bold}, and second-best are \underline{underlined}. 
		\vspace{-0.5cm}
	}
\end{table*}

\begin{table*}[!t]\footnotesize
	\setlength{\tabcolsep}{5pt}
	\centering
	\begin{tabular}{cc|cc|cc|cc|cc}
		\hline
		\multirow{3}*{\textbf{Category}} &  \multirow{3}*{\textbf{Ablation}} &  \multicolumn{8}{c}{\textbf{Dataset}}  \\
		\cline{3-10}
		~ & ~ & \multicolumn{2}{c|}{\textbf{Restaurant}} &  \multicolumn{2}{c|}{\textbf{Laptop}}  &  \multicolumn{2}{c|}{\textbf{MAMS}} &  \multicolumn{2}{c}{\textbf{Twitter}}\\
		\cline{3-10}
		~ & ~ & Acc.(\%)&F1.(\%) & Acc.(\%) &F1.(\%)  & Acc.(\%) &F1.(\%) & Acc.(\%) &F1.(\%) \\
		\hline 
		w/o AA & w/o syn. \& dep.(BERT+) & 84.99 & 78.51 & 79.11 & 75.76 & 82.71 & 82.22 & 75.48 & 74.54\\
		& w/o con. & 86.42 & 80.10 & 80.22 & 76.42 & 83.38 & 82.90 & 76.51 & 75.29 \\
		~ & w/o dep. & 86.60 & 81.51 & 81.80 & 78.48 & 84.58 & 84.09 & 76.81 & 75.86 \\
		\cline{2-10}
		~ & con.$\times$dep. & 86.86 & 80.82 & 80.85 & 77.27 & 84.21 & 83.76 & 76.51 & 75.37 \\
		~ & con.+dep. & 86.86 & 81.59 & 82.12 & 78.93 & 84.73 & 84.14 & \underline{77.40} & \underline{76.39} \\
		~ & con.$\oplus$dep. (BiSyn-GAT)& 87.49 & 81.63 & 82.44 & 79.15 & 84.90 & 84.43 & \textbf{77.99} & \textbf{76.80} \\
		\hline
		w/ AA  & con.+dep. & \underline{87.76} & \underline{82.18} & \underline{82.75} & \underline{79.16} & \underline{85.48} & \underline{85.05} & - & - \\
		~ & con.$\oplus$dep. (BiSyn-GAT+) & \textbf{87.94} & \textbf{82.43} & \textbf{82.91} & \textbf{79.38}  & \textbf{85.85} & \textbf{85.49} & - & - \\
		\hline
	\end{tabular}
	\vspace{-0.25cm}
	\caption{\label{Table:ablation-study} Ablation study. Notations ``con.'' and ``dep.'' represent syntax information from constituent tree and dependency tree, respectively. $\times,+,\oplus$ represent the position-wise dot, position-wise add, conditional position-wise add operations, respectively, when fusing two syntax information. ``AA'' represents modeling aspect-aspect relations. The best performances are in \textbf{bold}, and second-best are \underline{underlined}. 
		\vspace{-0.35cm}
	}
\end{table*}

\section{Experiment}
\subsection{Datasets and Setup}
We evaluate our models on four English dataset: Laptop, Restaurant datasets from SemEval2014 (Task 4) \cite{Pontiki2014SemEval2014T4}, MAMS \cite{jiang-etal-2019-challenge}, and Twitter \cite{dong-etal-2014-adaptive}. Laptop and Restaurant contain both multi-aspect and single-aspect sentences. Each sentence in MAMS contains at least two aspects with different sentiments. Twitter contains only one-aspect sentences. Dataset statistics are shown in Table~\ref{Table:dataset}. 

We adopt SuPar\footnote{https://github.com/yzhangcs/parser} as parser. Specifically, we use CRF constituency parser \cite{zhang-etal-2020-fast} to get the constituent tree; and following previous works~\cite{Wang2020RelationalGA,10.1109/TASLP.2020.3042009}, we use deep Biaffine Parser \cite{Dozat2017DeepBA} to get the dependency tree. Our context encoder is BERT-base-uncased \footnote{https://github.com/huggingface/transformers} model. Adam optimizer is adopted with a learning rate $2 \times 10^{-5}$ and a $L_2$ regulation $10^{-5}$ for model training. Number of GAT layers of one HGAT block is 3, and number of HGAT blocks is in range [1,3] on different datasets. ``Accuracy'' and ``Macro-Averaged F1'' are evaluation metrics. 
More details are in Appendix~\ref{sec:appendix_detail}.

\subsection{Baselines}
We compare our model with the following models:

1) Syntax-free baselines: \textbf{BERT-SPC} \cite{Song2019AttentionalEN}, \textbf{AEN-BERT} \cite{Song2019AttentionalEN};

2) Syntax-based baselines: \textbf{R-GAT} \cite{Wang2020RelationalGA}, \textbf{RGAT+} \cite{10.1109/TASLP.2020.3042009}, \textbf{DGEDT} \cite{tang-etal-2020-dependency}, \textbf{DualGCN} \cite{li-etal-2021-dual-graph};

3) Baselines that model aspect-aspect relations: \textbf{SDGCN-BERT}  \cite{ZHAO2020105443}, \textbf{InterGCN} \cite{liang-etal-2020-jointly};

Ours are also syntax-based, including:

a) \textbf{BiSyn-GAT+}: our full model, which contains the \emph{intra}-context module that combines two syntax information by \textbf{conditional position-wise add} operation, \emph{inter}-context module, and sentiment classifier to make predictions;

b) \textbf{BiSyn-GAT}: full model without \emph{inter}-context module;

Baselines and our models are all BERT-based.

\begin{table}[!t]\footnotesize
	\centering
	\setlength{\tabcolsep}{0.34pt}
	\renewcommand\arraystretch{0.95}
	\begin{tabular}{cc|cc|cc}
		\hline
		~ & \multirow{3}*{\textbf{Model}} &  \multicolumn{4}{c}{\textbf{Dataset}}  \\
		\cline{3-6}
		~ & ~ & \multicolumn{2}{c|}{\textbf{Restaurant}} &  \multicolumn{2}{c}{\textbf{MAMS}} \\
		~ & ~ & Acc.(\%)&F1.(\%) & Acc.(\%) &F1.(\%) \\
		\hline
		~ & BiSyn-GAT & 87.49 & 81.63  & 84.90 & 84.43 \\
		\hline
		aspect-context  & w/ Bi-relation & \textbf{87.94} & \textbf{82.43} & \textbf{85.85} & \textbf{85.49} \\
		graph & w/o Bi-relation & 87.85 & 82.27 & 85.10 & 84.69 \\
		\hline
		& adjacent  & 87.49  & 81.69 & 85.10 & 84.61 \\ 
		aspect graph& Bi-adjacent  & 87.40 & 81.53 & 85.18 & 84.74 \\
		& global & 87.49 &  81.70 & 85.32 & 84.88 \\
		\hline 
	\end{tabular}
	\vspace{-0.25cm}
	\caption{\label{Table:bi-exp-result} Performance comparison of aspect-context graph variants on Restaurant and MAMS dataset. The best performances are in \textbf{bold}.
		\vspace{-0.45cm}
	}
\end{table}

\begin{figure*}[!t]
	\centering
	\includegraphics[width=0.66\textwidth,height=0.34\textwidth]{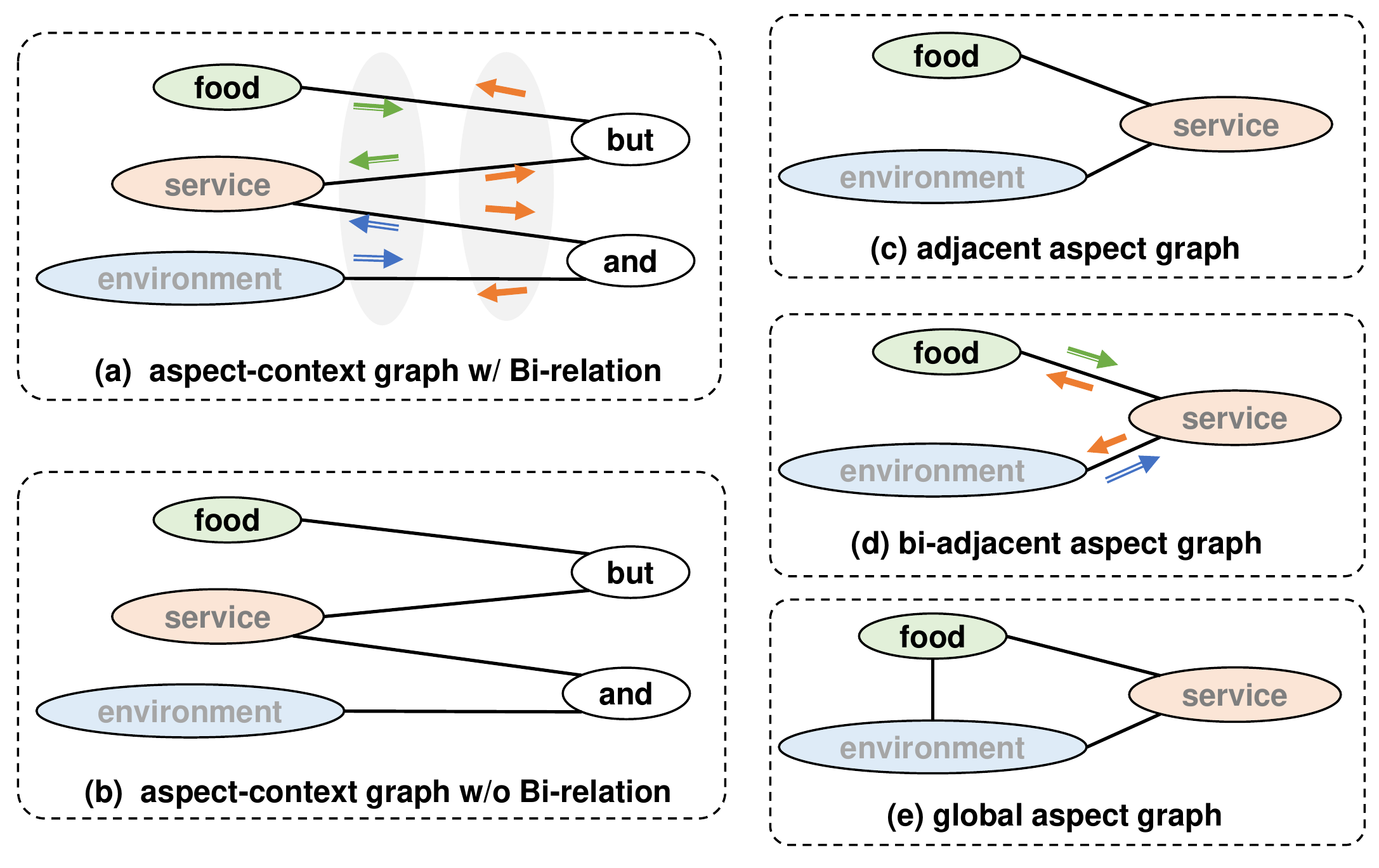}
	\vspace{-0.35cm}
	\caption{Illustrations of variants when investigating the effects of aspect-context graph.
	} 
	\label{Fig:graph_1}	
	\vspace{-0.6cm}
\end{figure*}

\subsection{Main Results}
Table \ref{Table:exp-result} shows results of the baselines and our models. For fairness of comparison, we present the reported results of those baselines.
Observations are: 
1) Our proposed models outperform most baselines, and our full model \textbf{\emph{BiSyn-GAT+}} achieves state-of-the-art performances in all datasets, especially 1.27 and 1.75 F1 improvements on Restaurant and MAMS. 
2) Models with syntax information outperform those without, which means syntax structure is helpful.
3) Our models show superiority to those that only use dependency information, which implies that constituent tree can provide profitable information.
4) \textbf{\emph{BiSyn-GAT+}} shows consistent improvement compared to \textbf{\emph{BiSyn-GAT}}, which means modeling aspect-aspect relations can improve performance, especially when more multi-aspect sentences are available, \eg 0.8 and 1.06 F1 improvements on Restaurant and MAMS.

\begin{table}[!t]\small
	\setlength{\tabcolsep}{0.95pt}
	\renewcommand\arraystretch{0.95}
	\centering
	\begin{tabular}{cc|cc|cc}
		\hline
		\multirow{2}*{\textbf{Model}} & \multirow{2}*{\textbf{Parser}} &  \multicolumn{2}{c|}{\textbf{Restaurant}} &  \multicolumn{2}{c}{\textbf{MAMS}}  \\
		~ & ~ & Acc.(\%)&F1.(\%) & Acc.(\%)&F1.(\%) \\
		\hline 
		\multicolumn{2}{c|}{Base} & 84.99 & 78.51 & 82.71 & 82.22 \\
		\hline
		\multirow{2}*{w/o dep.} &  Stanford Parser & 86.51 & 81.34 & 84.51 & 84.06\\
		& SuPar & 86.60 & 81.51 & 84.58 & 84.09 \\
		\hline
		\multirow{2}*{BiSyn-GAT} & Stanford Parser & 86.66 & 81.56 & 84.88 & 84.31 \\
		& SuPar & 87.49 & 81.63 & 84.90 & 84.43 \\
		\hline
		\multirow{2}*{BiSyn-GAT+} & Stanford Parser & 87.84 & 82.39 & 85.78 & 85.40 \\
		& SuPar & 87.94 & 82.43 & 85.85 & 85.49 \\
		
		\hline
	\end{tabular}
	\vspace{-0.25cm}
	\caption{\label{Table:parser_effects} Experiments results with different parsers. w/o dep. is one variant of BiSyn-GAT, only using constituent information.}
	\vspace{-0.55cm}
\end{table}

\begin{table*}[!t]\footnotesize
	\renewcommand\arraystretch{0.90}
	\centering
	\begin{tabular}{c|c|c|c}
		\hline
		\multicolumn{1}{c|}{Sentences} & Aspects & BiSyn-GAT & BiSyn-GAT+ \\
		\hline
		it doesn't look like much on the \textbf{outside}$\mathbf{_{neg}}$ \underline{, but} the minute &  outside & neu \XSolidBrush & neg \Checkmark \\
		you walk inside, it's a whole other \textbf{atmosphere}$\mathbf{_{pos}}$. & atmosphere & pos \Checkmark  & pos \Checkmark \\
		\hline
		while the \textbf{service}$\mathbf{_{neg}}$ \underline{and} \textbf{setting}$\mathbf{_{neg}}$ were average & service & neg \Checkmark & neg \Checkmark \\
		\underline{,} the \textbf{food}$\mathbf{_{pos}}$ was excellent.  & setting& neu \XSolidBrush  & neg \Checkmark \\
		& food & pos \Checkmark & pos \Checkmark \\
		\hline
		food was average, the \textbf{appetizers}$\mathbf{_{pos}}$ \underline{were} & appetizers & pos \Checkmark & pos \Checkmark \\
		\underline{better than the} \textbf{main courses}$\mathbf{_{neu}}$.& main courses & pos \XSolidBrush & neu \Checkmark \\
		\hline
		i have no complaints about the \textbf{wait}$\mathbf{_{pos}}$ \underline{or} the \textbf{service}$\mathbf{_{pos}}$  & wait & neu \XSolidBrush & pos \Checkmark \\
		\underline{but} the \textbf{pizza}$\mathbf{_{neg}}$ was bit at all something to write home about.& service & neg \XSolidBrush & pos \Checkmark\\
		& pizza & neg \Checkmark & neg \Checkmark \\
		\hline
	\end{tabular}
	\vspace{-0.25cm}
	\caption{\label{Table:comparsion} Predictions from \emph{BiSyn-GAT} and \emph{BiSyn-GAT+}. The notations pos, neg, and neu in the table represent positive, negative, and neutral. For each sentence, the aspects are displayed in bold, with golden sentiment polarities as the subscripts. The phrase segmentation words are shown underline between the corresponding two aspects. False predictions are marked with \XSolidBrush  while true predictions are marked with \Checkmark.}
	\vspace{-0.5cm}
\end{table*}

\subsection{Ablation Study} 
We also conduct an ablation study to verify the effectiveness of our proposed method. The results are shown in Table \ref{Table:ablation-study}. We set the context encoder of our model as the base model, \ie \emph{BERT+}. The observations are that:
1) \emph{BERT+} achieves the lowest performance, which shows syntax information is helpful in ABSA task. 
2) In category \emph{w/o AA}, \textbf{w/o con.} is inferior to \textbf{w/o dep.}, which means syntax information of Con.Tree is useful. Moreover, the comparison between \textbf{w/o con.} and \textbf{con.$\times$dep.} verifies that some dependency edges that cross the phrases indeed bring noise, as the former considers all dependency edges and the latter ignores those across phrases obtained from Con.Tree for each layer.
3) Fusing two syntax information in the proper ways can boost performance. In category \emph{w/o AA}, \textbf{con.+dep.} and \textbf{con.$\oplus$dep.} both outperform \textbf{w/o dep.} and \textbf{w/o con.} in all datasets. However, \textbf{con.$\times$dep.} is inferior to \textbf{w/o dep.}. One possible reason is that the position-wise dot operation ignores most connections within phrases, causing the graphs to be more sparse. It also verifies that words within the same phrases of Con.Tree are essential for aligning aspects and corresponding opinions.
4) Modeling aspect-aspect relations is beneficial from the comparison between \emph{w/ AA} and \emph{w/o AA}, especially in Restaurant and MAMS that contain more multi-aspect sentences.  

\subsection{Effects of Aspect-context graph\label{sec:aspect-context-graph}}
We also investigate the effects of our bi-relational modeling of the proposed aspect-context graph. Firstly, we use \textbf{BiSyn-GAT} as base model to see whether the approach modeling aspects relations improves the performance;
Secondly, based on our proposed aspect-context graph, we consider two variants: (a) \textbf{w/ Bi-relation}, a directed one that distinguishes the influence one aspect imposes on other aspects and is received from other aspects, {\ie} our full model BiSyn-GAT+; (b) \textbf{w/o Bi-relation}, an undirected one that ignores the direction of the influence;
Thirdly, inspired by \citet{ZHAO2020105443}, we define the aspect graph as the graph with all aspects as its nodes, \ie our aspect-context graph without any segmentation terms. Based on the aspect graph, we propose three variants: (c) \textbf{adjacent aspect graph}, an undirected one where neighbor aspects are connected; (d) \textbf{bi-adjacent aspect graph}, a directed one where neighbor aspects are connected; (e) \textbf{global aspect graph}, an undirected one where all aspects are connected;
The above five variants are illustrated in Figure~\ref{Fig:graph_1}. Experimental results are shown in Table~\ref{Table:bi-exp-result} and we can observe that: 1) w/ Bi-relation (\ie BiSyn-GAT+) outperforms w/o Bi-relation consistently, which indicates distinguishing the bi-relational influences is beneficial; 2) Overall, aspect-context graph shows superiority compared with aspect graph, which means the phrase segmentation terms can help model aspects relations; 3) Unlike in aspect-context graph, bi-adjacent aspect graph does not guarantee performance improvement compared with adjacent aspect graph, which reflects the importance of phrase segmentation terms when modeling aspect-aspect relations;
4) Overall, global aspect graph performs better than adjacent aspect graph, which is correlated with the results in \citet{ZHAO2020105443}; 5) In Restaurant dataset, adjacent aspect graph and global aspect graph show comparable performance. One possible reason is that the number of samples that contain at least three aspects is very limited, as shown in Table~\ref{Table:dataset-detail_multi_aspect} of Appendix. And adjacent aspect graph equals global aspect graph when faced with two aspects.

\subsection{Effects of Parsing}
We conduct experiments to study the influence of paring accuracy on model performance. Two parsers are selected:~(a) Stanford Parser~\cite{manning-etal-2014-stanford}, a well-known toolkit; it has transition-based dependency parser~\cite{Chen2014AFA} and shift-reduce constituency parser~\cite{zhu-etal-2013-fast}; (b) SuPar, which RGAT+ ~\cite{10.1109/TASLP.2020.3042009} and our proposed models adopt; it has deep biaffine dependency parser~\cite{Dozat2017DeepBA} and neural CRF constituency parser~\cite{zhang-etal-2020-fast}. Generally, SuPar has better parsing performances than Stanford Parser. We use BERT+ as the base model and compare the performance of model \textbf{w/o dep}, \textbf{Bisyn-GAT}, \textbf{BiSyn-GAT+} when using different parsers. The results are shown in Table~\ref{Table:parser_effects}. Observations are that: 1) With Stanford Parser, our models can also achieve good performance. 2) Models with SuPar perform better than models with Stanford Parser, which is correlated with the parsing accuracy of two parsers.

\subsection{Case Study}
As shown in Figure~\ref{Table:comparsion}, we present four examples to help better understand our proposed model, especially \emph{inter}-context module when faced with complex sentences. 
The first is a comparative sentence with two clauses connected by the conjunction ``but''. Both models make correct predictions for \textbf{atmosphere}. However, \emph{BiSyn-GAT} predicts wrong over \textbf{outside} while \emph{BiSyn-GAT+} still makes a correct prediction, which show the \emph{inter}-context module correctly captures the reversed sentiment relation between \textbf{outside} and \textbf{atmosphere} by phrase segmentation terms ``, but''. The rest examples all show that \emph{inter}-context module can use relations across aspects to help correct the predictions.

\section{Conclusion}
In this paper, we propose the BiSyn-GAT+ framework to model the sentiment-aware context of each aspect and sentiment relations across aspects for learning by fully exploiting the syntax information of the constituent tree. It includes two well-designed modules: 1) \emph{intra}-context module that fuses related semantic and syntax information hierarchically; 2) \emph{inter}-context module that models relations across aspects with the constructed aspect-context graph. To the best of our knowledge, it is the first work to exploit the constituent tree with GNNs for the ABSA task. Moreover, our proposed model achieves state-of-the-art performances on four benchmark datasets.


\section*{Acknowledgements}
This work was supported in part by the National Natural Science Foundation of China under Grant No.61602197, Grant No.L1924068, Grant No.61772076, in part by CCF-AFSG Research Fund under Grant No.RF20210005, and in part by the fund of Joint Laboratory of HUST and Pingan Property \& Casualty Research (HPL). The authors would also like to thank the anonymous reviewers for their comments on improving the quality of this paper.

\bibliography{anthology,custom}

\begin{thebibliography}{37}
\expandafter\ifx\csname natexlab\endcsname\relax\def\natexlab#1{#1}\fi

\bibitem[{Bai et~al.(2020)Bai, Liu, and Zhang}]{10.1109/TASLP.2020.3042009}
Xuefeng Bai, Pengbo Liu, and Yue Zhang. 2020.
\newblock Investigating typed syntactic dependencies for targeted sentiment
  classification using graph attention neural network.
\newblock \emph{IEEE/ACM Transactions on Audio, Speech, and Language
  Processing}, 29:503--514.

\bibitem[{Chen et~al.(2020)Chen, Teng, and Zhang}]{Chen2020InducingTL}
Chenhua Chen, Zhiyang Teng, and Yue Zhang. 2020.
\newblock \href {https://doi.org/10.18653/v1/2020.emnlp-main.451} {Inducing
  target-specific latent structures for aspect sentiment classification}.
\newblock In \emph{Proceedings of the 2020 Conference on Empirical Methods in
  Natural Language Processing (EMNLP)}, pages 5596--5607, Online. Association
  for Computational Linguistics.

\bibitem[{Chen and Manning(2014)}]{Chen2014AFA}
Danqi Chen and Christopher Manning. 2014.
\newblock \href {https://doi.org/10.3115/v1/D14-1082} {A fast and accurate
  dependency parser using neural networks}.
\newblock In \emph{Proceedings of the 2014 Conference on Empirical Methods in
  Natural Language Processing ({EMNLP})}, pages 740--750, Doha, Qatar.
  Association for Computational Linguistics.

\bibitem[{Chen et~al.(2017)Chen, Sun, Bing, and
  Yang}]{chen-etal-2017-recurrent}
Peng Chen, Zhongqian Sun, Lidong Bing, and Wei Yang. 2017.
\newblock \href {https://doi.org/10.18653/v1/D17-1047} {Recurrent attention
  network on memory for aspect sentiment analysis}.
\newblock In \emph{Proceedings of the 2017 Conference on Empirical Methods in
  Natural Language Processing}, pages 452--461, Copenhagen, Denmark.
  Association for Computational Linguistics.

\bibitem[{Devlin et~al.(2019)Devlin, Chang, Lee, and
  Toutanova}]{Devlin2019BERTPO}
Jacob Devlin, Ming-Wei Chang, Kenton Lee, and Kristina Toutanova. 2019.
\newblock \href {https://doi.org/10.18653/v1/N19-1423} {{BERT}: Pre-training of
  deep bidirectional transformers for language understanding}.
\newblock In \emph{Proceedings of the 2019 Conference of the North {A}merican
  Chapter of the Association for Computational Linguistics: Human Language
  Technologies, Volume 1 (Long and Short Papers)}, pages 4171--4186,
  Minneapolis, Minnesota. Association for Computational Linguistics.

\bibitem[{Dong et~al.(2014)Dong, Wei, Tan, Tang, Zhou, and
  Xu}]{dong-etal-2014-adaptive}
Li~Dong, Furu Wei, Chuanqi Tan, Duyu Tang, Ming Zhou, and Ke~Xu. 2014.
\newblock \href {https://doi.org/10.3115/v1/P14-2009} {Adaptive recursive
  neural network for target-dependent {T}witter sentiment classification}.
\newblock In \emph{Proceedings of the 52nd Annual Meeting of the Association
  for Computational Linguistics (Volume 2: Short Papers)}, pages 49--54,
  Baltimore, Maryland. Association for Computational Linguistics.

\bibitem[{Dozat and Manning(2017)}]{Dozat2017DeepBA}
Timothy Dozat and Christopher~D. Manning. 2017.
\newblock \href {https://openreview.net/forum?id=Hk95PK9le} {Deep biaffine
  attention for neural dependency parsing}.
\newblock In \emph{5th International Conference on Learning Representations,
  {ICLR} 2017, Toulon, France, April 24-26, 2017, Conference Track
  Proceedings}. OpenReview.net.

\bibitem[{Fan et~al.(2018)Fan, Feng, and Zhao}]{fan-etal-2018-multi}
Feifan Fan, Yansong Feng, and Dongyan Zhao. 2018.
\newblock \href {https://doi.org/10.18653/v1/D18-1380} {Multi-grained attention
  network for aspect-level sentiment classification}.
\newblock In \emph{Proceedings of the 2018 Conference on Empirical Methods in
  Natural Language Processing}, pages 3433--3442, Brussels, Belgium.
  Association for Computational Linguistics.

\bibitem[{Hazarika et~al.(2018)Hazarika, Poria, Vij, Krishnamurthy, Cambria,
  and Zimmermann}]{hazarika2018modeling}
Devamanyu Hazarika, Soujanya Poria, Prateek Vij, Gangeshwar Krishnamurthy, Erik
  Cambria, and Roger Zimmermann. 2018.
\newblock \href {https://doi.org/10.18653/v1/N18-2043} {Modeling inter-aspect
  dependencies for aspect-based sentiment analysis}.
\newblock In \emph{Proceedings of the 2018 Conference of the North {A}merican
  Chapter of the Association for Computational Linguistics: Human Language
  Technologies, Volume 2 (Short Papers)}, pages 266--270, New Orleans,
  Louisiana. Association for Computational Linguistics.

\bibitem[{Hu et~al.(2019)Hu, Zhao, Zhang, Cai, Su, Cheng, and
  Shen}]{hu-etal-2019-constrained}
Mengting Hu, Shiwan Zhao, Li~Zhang, Keke Cai, Zhong Su, Renhong Cheng, and
  Xiaowei Shen. 2019.
\newblock \href {https://doi.org/10.18653/v1/D19-1467} {{CAN}: Constrained
  attention networks for multi-aspect sentiment analysis}.
\newblock In \emph{Proceedings of the 2019 Conference on Empirical Methods in
  Natural Language Processing and the 9th International Joint Conference on
  Natural Language Processing (EMNLP-IJCNLP)}, pages 4601--4610, Hong Kong,
  China. Association for Computational Linguistics.

\bibitem[{Jiang et~al.(2019)Jiang, Chen, Xu, Ao, and
  Yang}]{jiang-etal-2019-challenge}
Qingnan Jiang, Lei Chen, Ruifeng Xu, Xiang Ao, and Min Yang. 2019.
\newblock \href {https://doi.org/10.18653/v1/D19-1654} {A challenge dataset and
  effective models for aspect-based sentiment analysis}.
\newblock In \emph{Proceedings of the 2019 Conference on Empirical Methods in
  Natural Language Processing and the 9th International Joint Conference on
  Natural Language Processing (EMNLP-IJCNLP)}, pages 6280--6285, Hong Kong,
  China. Association for Computational Linguistics.

\bibitem[{Lan et~al.(2020)Lan, Mao, Wei, Gao, and Huang}]{10.1145/3423168}
Tian Lan, Xian-Ling Mao, Wei Wei, Xiaoyan Gao, and Heyan Huang. 2020.
\newblock \href {https://doi.org/10.1145/3423168} {Pone: A novel automatic
  evaluation metric for open-domain generative dialogue systems}.
\newblock \emph{ACM Trans. Inf. Syst.}, 39(1).

\bibitem[{Li et~al.(2020{\natexlab{a}})Li, Chen, Ren, Ren, Tu, and
  Chen}]{li-etal-2020-empdg}
Qintong Li, Hongshen Chen, Zhaochun Ren, Pengjie Ren, Zhaopeng Tu, and Zhumin
  Chen. 2020{\natexlab{a}}.
\newblock \href {https://doi.org/10.18653/v1/2020.coling-main.394} {{E}mp{DG}:
  Multi-resolution interactive empathetic dialogue generation}.
\newblock In \emph{Proceedings of the 28th International Conference on
  Computational Linguistics}, pages 4454--4466, Barcelona, Spain (Online).
  International Committee on Computational Linguistics.

\bibitem[{Li et~al.(2021)Li, Chen, Feng, Ma, Wang, and
  Hovy}]{li-etal-2021-dual-graph}
Ruifan Li, Hao Chen, Fangxiang Feng, Zhanyu Ma, Xiaojie Wang, and Eduard Hovy.
  2021.
\newblock \href {https://doi.org/10.18653/v1/2021.acl-long.494} {Dual graph
  convolutional networks for aspect-based sentiment analysis}.
\newblock In \emph{Proceedings of the 59th Annual Meeting of the Association
  for Computational Linguistics and the 11th International Joint Conference on
  Natural Language Processing (Volume 1: Long Papers)}, pages 6319--6329,
  Online. Association for Computational Linguistics.

\bibitem[{Li et~al.(2020{\natexlab{b}})Li, Yin, and hua
  Zhong}]{Li2020SentenceCA}
Yuncong Li, Cunxiang Yin, and Sheng hua Zhong. 2020{\natexlab{b}}.
\newblock Sentence constituent-aware aspect-category sentiment analysis with
  graph attention networks.
\newblock In \emph{NLPCC}.

\bibitem[{Li et~al.(2019)Li, Wei, Zhang, Zhang, and Li}]{Li2019ExploitingCT}
Zheng Li, Ying Wei, Yu~Zhang, Xiang Zhang, and Xin Li. 2019.
\newblock \href {https://doi.org/10.1609/aaai.v33i01.33014253} {Exploiting
  coarse-to-fine task transfer for aspect-level sentiment classification}.
\newblock In \emph{The Thirty-Third {AAAI} Conference on Artificial
  Intelligence, {AAAI} 2019, The Thirty-First Innovative Applications of
  Artificial Intelligence Conference, {IAAI} 2019, The Ninth {AAAI} Symposium
  on Educational Advances in Artificial Intelligence, {EAAI} 2019, Honolulu,
  Hawaii, USA, January 27 - February 1, 2019}, pages 4253--4260. {AAAI} Press.

\bibitem[{Liang et~al.(2020)Liang, Yin, Gui, Du, and
  Xu}]{liang-etal-2020-jointly}
Bin Liang, Rongdi Yin, Lin Gui, Jiachen Du, and Ruifeng Xu. 2020.
\newblock \href {https://doi.org/10.18653/v1/2020.coling-main.13} {Jointly
  learning aspect-focused and inter-aspect relations with graph convolutional
  networks for aspect sentiment analysis}.
\newblock In \emph{Proceedings of the 28th International Conference on
  Computational Linguistics}, pages 150--161, Barcelona, Spain (Online).
  International Committee on Computational Linguistics.

\bibitem[{Majumder et~al.(2018)Majumder, Poria, Gelbukh, Akhtar, Cambria, and
  Ekbal}]{majumder-etal-2018-iarm}
Navonil Majumder, Soujanya Poria, Alexander Gelbukh, Md.~Shad Akhtar, Erik
  Cambria, and Asif Ekbal. 2018.
\newblock \href {https://doi.org/10.18653/v1/D18-1377} {{IARM}: Inter-aspect
  relation modeling with memory networks in aspect-based sentiment analysis}.
\newblock In \emph{Proceedings of the 2018 Conference on Empirical Methods in
  Natural Language Processing}, pages 3402--3411, Brussels, Belgium.
  Association for Computational Linguistics.

\bibitem[{Manning et~al.(2014)Manning, Surdeanu, Bauer, Finkel, Bethard, and
  McClosky}]{manning-etal-2014-stanford}
Christopher Manning, Mihai Surdeanu, John Bauer, Jenny Finkel, Steven Bethard,
  and David McClosky. 2014.
\newblock \href {https://doi.org/10.3115/v1/P14-5010} {The {S}tanford
  {C}ore{NLP} natural language processing toolkit}.
\newblock In \emph{Proceedings of 52nd Annual Meeting of the Association for
  Computational Linguistics: System Demonstrations}, pages 55--60, Baltimore,
  Maryland. Association for Computational Linguistics.

\bibitem[{Pontiki et~al.(2014)Pontiki, Galanis, Pavlopoulos, Papageorgiou,
  Androutsopoulos, and Manandhar}]{Pontiki2014SemEval2014T4}
Maria Pontiki, Dimitris Galanis, John Pavlopoulos, Harris Papageorgiou, Ion
  Androutsopoulos, and Suresh Manandhar. 2014.
\newblock \href {https://doi.org/10.3115/v1/S14-2004} {{S}em{E}val-2014 task 4:
  Aspect based sentiment analysis}.
\newblock In \emph{Proceedings of the 8th International Workshop on Semantic
  Evaluation ({S}em{E}val 2014)}, pages 27--35, Dublin, Ireland. Association
  for Computational Linguistics.

\bibitem[{Qiu et~al.(2021)Qiu, Huang, Chen, Ji, Qu, Wei, Huang, and
  Zhang}]{Qiu2021ReinforcedHB}
Minghui Qiu, Xinjing Huang, Cen-Chieh Chen, Feng Ji, Chen Qu, Wei Wei, Jun
  Huang, and Yin Zhang. 2021.
\newblock Reinforced history backtracking for conversational question
  answering.
\newblock In \emph{AAAI}.

\bibitem[{Song et~al.(2019)Song, Wang, Jiang, Liu, and
  Rao}]{Song2019AttentionalEN}
Youwei Song, Jiahai Wang, Tao Jiang, Zhiyue Liu, and Yanghui Rao. 2019.
\newblock Attentional encoder network for targeted sentiment classification.
\newblock \emph{ArXiv}, abs/1902.09314.

\bibitem[{Tang et~al.(2020)Tang, Ji, Li, and Zhou}]{tang-etal-2020-dependency}
Hao Tang, Donghong Ji, Chenliang Li, and Qiji Zhou. 2020.
\newblock \href {https://doi.org/10.18653/v1/2020.acl-main.588} {Dependency
  graph enhanced dual-transformer structure for aspect-based sentiment
  classification}.
\newblock In \emph{Proceedings of the 58th Annual Meeting of the Association
  for Computational Linguistics}, pages 6578--6588, Online. Association for
  Computational Linguistics.

\bibitem[{Wang et~al.(2020{\natexlab{a}})Wang, Shen, Yang, Quan, and
  Wang}]{Wang2020RelationalGA}
Kai Wang, Weizhou Shen, Yunyi Yang, Xiaojun Quan, and Rui Wang.
  2020{\natexlab{a}}.
\newblock \href {https://doi.org/10.18653/v1/2020.acl-main.295} {Relational
  graph attention network for aspect-based sentiment analysis}.
\newblock In \emph{Proceedings of the 58th Annual Meeting of the Association
  for Computational Linguistics}, pages 3229--3238, Online. Association for
  Computational Linguistics.

\bibitem[{Wang et~al.(2016)Wang, Huang, Zhu, and
  Zhao}]{wang-etal-2016-attention}
Yequan Wang, Minlie Huang, Xiaoyan Zhu, and Li~Zhao. 2016.
\newblock \href {https://doi.org/10.18653/v1/D16-1058} {Attention-based {LSTM}
  for aspect-level sentiment classification}.
\newblock In \emph{Proceedings of the 2016 Conference on Empirical Methods in
  Natural Language Processing}, pages 606--615, Austin, Texas. Association for
  Computational Linguistics.

\bibitem[{Wang et~al.(2020{\natexlab{b}})Wang, Wei, Cong, Li, Mao, and
  Qiu}]{wang2020global}
Ziyang Wang, Wei Wei, Gao Cong, Xiao{-}Li Li, Xianling Mao, and Minghui Qiu.
  2020{\natexlab{b}}.
\newblock \href {https://doi.org/10.1145/3397271.3401142} {Global context
  enhanced graph neural networks for session-based recommendation}.
\newblock In \emph{Proceedings of the 43rd International {ACM} {SIGIR}
  conference on research and development in Information Retrieval, {SIGIR}
  2020, Virtual Event, China, July 25-30, 2020}, pages 169--178. {ACM}.

\bibitem[{Wei et~al.(2011)Wei, Cong, Li, Ng, and Li}]{Wei2011IntegratingCQ}
Wei Wei, Gao Cong, Xiaoli Li, See{-}Kiong Ng, and Guohui Li. 2011.
\newblock \href {http://www.aaai.org/ocs/index.php/AAAI/AAAI11/paper/view/3626}
  {Integrating community question and answer archives}.
\newblock In \emph{Proceedings of the Twenty-Fifth {AAAI} Conference on
  Artificial Intelligence, {AAAI} 2011, San Francisco, California, USA, August
  7-11, 2011}. {AAAI} Press.

\bibitem[{Wei et~al.(2019)Wei, Liu, Mao, Guo, Zhu, Zhou, and
  Hu}]{wei2019emotion}
Wei Wei, Jiayi Liu, Xianling Mao, Guibing Guo, Feida Zhu, Pan Zhou, and Yuchong
  Hu. 2019.
\newblock \href {https://doi.org/10.1145/3357384.3357937} {Emotion-aware chat
  machine: Automatic emotional response generation for human-like emotional
  interaction}.
\newblock In \emph{Proceedings of the 28th {ACM} International Conference on
  Information and Knowledge Management, {CIKM} 2019, Beijing, China, November
  3-7, 2019}, pages 1401--1410. {ACM}.

\bibitem[{Wei et~al.(2021)Wei, Liu, Mao, Guo, Zhu, Zhou, Hu, and
  Feng}]{wei2021target}
Wei Wei, Jiayi Liu, Xianling Mao, Guibing Guo, Feida Zhu, Pan Zhou, Yuchong Hu,
  and Shanshan Feng. 2021.
\newblock Target-guided emotion-aware chat machine.
\newblock \emph{ACM Transactions on Information Systems (TOIS)}, 39(4):1--24.

\bibitem[{Yang et~al.(2018)Yang, Yang, Wang, and Xie}]{Yang2018MultiEntityAS}
Jun Yang, Runqi Yang, Chongjun Wang, and Junyuan Xie. 2018.
\newblock \href
  {https://www.aaai.org/ocs/index.php/AAAI/AAAI18/paper/view/17036}
  {Multi-entity aspect-based sentiment analysis with context, entity and aspect
  memory}.
\newblock In \emph{Proceedings of the Thirty-Second {AAAI} Conference on
  Artificial Intelligence, (AAAI-18), the 30th innovative Applications of
  Artificial Intelligence (IAAI-18), and the 8th {AAAI} Symposium on
  Educational Advances in Artificial Intelligence (EAAI-18), New Orleans,
  Louisiana, USA, February 2-7, 2018}, pages 6029--6036. {AAAI} Press.

\bibitem[{Yang et~al.(2020)Yang, Xu, and Gao}]{yang2020cm}
Kaicheng Yang, Hua Xu, and Kai Gao. 2020.
\newblock \href {https://doi.org/10.1145/3394171.3413690} {{CM-BERT:}
  cross-modal {BERT} for text-audio sentiment analysis}.
\newblock In \emph{{MM} '20: The 28th {ACM} International Conference on
  Multimedia, Virtual Event / Seattle, WA, USA, October 12-16, 2020}, pages
  521--528.

\bibitem[{Zhang et~al.(2019)Zhang, Li, and Song}]{zhang-etal-2019-aspect}
Chen Zhang, Qiuchi Li, and Dawei Song. 2019.
\newblock \href {https://doi.org/10.18653/v1/D19-1464} {Aspect-based sentiment
  classification with aspect-specific graph convolutional networks}.
\newblock In \emph{Proceedings of the 2019 Conference on Empirical Methods in
  Natural Language Processing and the 9th International Joint Conference on
  Natural Language Processing (EMNLP-IJCNLP)}, pages 4568--4578, Hong Kong,
  China. Association for Computational Linguistics.

\bibitem[{Zhang et~al.(2018)Zhang, Wang, and Liu}]{zhang2018deep}
Lei Zhang, Shuai Wang, and Bing Liu. 2018.
\newblock Deep learning for sentiment analysis: A survey.
\newblock \emph{Wiley Interdisciplinary Reviews: Data Mining and Knowledge
  Discovery}, 8(4):e1253.

\bibitem[{Zhang et~al.(2020)Zhang, Zhou, and Li}]{zhang-etal-2020-fast}
Yu~Zhang, Houquan Zhou, and Zhenghua Li. 2020.
\newblock \href {https://doi.org/10.24963/ijcai.2020/560} {Fast and accurate
  neural {CRF} constituency parsing}.
\newblock In \emph{Proceedings of the Twenty-Ninth International Joint
  Conference on Artificial Intelligence, {IJCAI} 2020}, pages 4046--4053.
  ijcai.org.

\bibitem[{Zhao et~al.(2020)Zhao, Hou, and Wu}]{ZHAO2020105443}
Pinlong Zhao, Linlin Hou, and Ou~Wu. 2020.
\newblock Modeling sentiment dependencies with graph convolutional networks for
  aspect-level sentiment classification.
\newblock \emph{Knowledge-Based Systems}, 193:105443.

\bibitem[{Zhao et~al.(2022)Zhao, Wei, Zou, and Mao}]{zhao2022multi}
Sen Zhao, Wei Wei, Ding Zou, and Xianling Mao. 2022.
\newblock \href {https://arxiv.org/abs/2202.11425} {Multi-view intent
  disentangle graph networks for bundle recommendation}.
\newblock \emph{ArXiv preprint}, abs/2202.11425.

\bibitem[{Zhu et~al.(2013)Zhu, Zhang, Chen, Zhang, and
  Zhu}]{zhu-etal-2013-fast}
Muhua Zhu, Yue Zhang, Wenliang Chen, Min Zhang, and Jingbo Zhu. 2013.
\newblock \href {https://aclanthology.org/P13-1043} {Fast and accurate
  shift-reduce constituent parsing}.
\newblock In \emph{Proceedings of the 51st Annual Meeting of the Association
  for Computational Linguistics (Volume 1: Long Papers)}, pages 434--443,
  Sofia, Bulgaria. Association for Computational Linguistics.

\end{thebibliography}
\bibliographystyle{acl_natbib}

\appendix

\clearpage
\section{Dataset and Implementation Detail\label{sec:appendix_detail}}
		\subsection{Statistics of constituent tree depth}
		\begin{table}[!t]\footnotesize
		\centering
		\setlength{\tabcolsep}{1.7pt}
		\begin{tabular}{c|cc|cc|ccc|cc}
			\hline
			\textbf{Con.} & \multicolumn{9}{c}{\textbf{Dataset}}  \\
			\cline{2-10}
			\textbf{Tree} & \multicolumn{2}{c|}{\textbf{Restaurant}} &  \multicolumn{2}{c|}{\textbf{Laptop}}  &  \multicolumn{3}{c|}{\textbf{MAMS}} &  \multicolumn{2}{c}{\textbf{Twitter}}\\
			\textbf{Depth} & Train & Test & Train & Test & Train & Valid & Test & Train & Test \\
			\hline 
			1 & 177 & 68 & 206 & 84 & 208 & 16 & 19 & 1215 & 117 \\
			\hline
			2 & 369 & 135 & 724 & 247 & 1301 & 152 & 141 & 1066 & \textbf{147} \\
			\hline
			3 & \textbf{462} & \textbf{148} & \textbf{936} & \textbf{312} & \textbf{2265} & 244 & 261 & \textbf{1186} & 123 \\
			\hline
			4 & 363 & 108 & 612 & 202 & 2085 & \textbf{276} & \textbf{292} & 947 & 96 \\
			\hline
			5 & 311 & 75 & 429 & 116 & 1761 & 203 & 194 & 677 & 79 \\
			\hline
			6 & 237 & 40 & 266 & 73 & 1211 & 141 & 157 & 414 & 57 \\
			\hline
			7 & 136 & 27 & 205 & 41 & 901 & 99 & 117 & 246 & 23 \\
			\hline
			8 & 108 & 10 & 106 & 18 & 545 & 81 & 65 & 145 & 22 \\
			\hline
			9 & 59 & 8 & 43 & 14 & 380 & 57 & 34 & 86 & 8 \\
			\hline
			$\geq 10$ & 60 & 13 & 81 & 12 & 529 & 63 & 56 & 69 & 5 \\
			\hline
			MAX. & 18 & 13 & 17 & 13 & 19 & 17 & 15 & 14 & 11 \\
			\hline
		\end{tabular}
		\vspace{-0.3cm}
		\caption{\label{Table:dataset-detail_con_depth} Depth distribution of parsed constituent trees on four datasets. The maximums are in \textbf{bold}. The last row lists the max tree depth of each dataset.
			\vspace{-0.2cm}
		}
	\end{table}
	\begin{table}[!t]\footnotesize
		\centering
		\setlength{\tabcolsep}{1.7pt}
		\begin{tabular}{c|cc|cc|ccc}
			\hline
			\textbf{Multi.} & \multicolumn{7}{c}{\textbf{Dataset}}  \\
			\cline{2-8}
			\textbf{Aspect} & \multicolumn{2}{c|}{\textbf{Restaurant}} &  \multicolumn{2}{c|}{\textbf{Laptop}}  &  \multicolumn{3}{c}{\textbf{MAMS}} \\
			\textbf{Distribution} & Train & Test & Train & Test & Train & Valid & Test \\
			\hline 
			2 & 555 & 192 & 343 & 101 & 2568 & 285 & 264  \\
			\hline
			3 & 261 & 73 & 137 & 33 & 1169 & 136 & 173 \\
			\hline
			4 & 103 & 31 & 40 & 9 & 364 & 55 & 45 \\
			\hline
			5 & 32 & 14 & 9 & 6 & 126 & 16 & 10 \\
			\hline
			6 & 11 & 3 & 5 & 1 & 48 & 5 & 5  \\
			\hline
			7 & 5 & 1 & 3 & - & 13 & 2 & -  \\
			\hline
			8 & 3 & - & - & - & 6 & - & 1  \\
			\hline
			9 & 1 & - & - & - & 1 & - & -  \\
			\hline
			10 & - & - & - & - & 1 & 1 & 1  \\
			\hline
			11 & - & - & - & - & 1 & 1 & 1  \\
			\hline
			13 & - & 1 & 1 & - & - & - \\
			\hline
		\end{tabular}
		\vspace{-0.3cm}
		\caption{\label{Table:dataset-detail_multi_aspect} Multi.aspect distribution of three datasets.
			\vspace{-0.5cm}
		}
	\end{table}
	
	Table~\ref{Table:dataset-detail_con_depth} shows more detailed statistics about four benchmark datasets at the aspect level. We define the ``constituent tree depth'' as the number of nodes in the path from the aspect term node to the root node in the Con.Tree. It means we treat the layer that the aspect term is in as the bottom layer for constituent graph construction and drop layers below it. The aspect term has no other neighbors in those layers and thus fails to update its representation through the graph encoder. According to the constituent tree depth statistics, we set the number of GAT layers of one HGAT block in the syntax encoder to 3, the most common depth.
	
	
	\subsection{Multi-aspect Distribution of datasets}
	Table~\ref{Table:dataset-detail_multi_aspect} shows the multi-aspect distribution of the Restaurant, Laptop, and MAMS datasets. This can explain the improvement of BiSyn-GAT+ compared to BiSyn-GAT on different datasets: MAMS > Restaurant > Laptop. MAMS contains the most multi-aspect sentences that our proposed \emph{Inter-context} module can fully utilize.
	
	\subsection{Training Detail}
	 The numbers of parameters of BiSyn-GAT and BiSyn-GAT+ are 112M and 233M. Each epoch takes about 60s or 70s in RTX 2080 Ti.
	 We test the model that performs best on validation data, and for datasets without official validation data, we follow the dataset settings of previous work ~\cite{10.1109/TASLP.2020.3042009}. We use the grid search to find the best parameters for our model and report the maximum results. The number of HGAT blocks within our relation encoder is in range [1,3] on different datasets and the number of its inner GAT layers is set to 2; the dropout rate is 0.1 for the input and output and is in the range [0.2, 0.7] between layers; In each HGAT block of our syntax encoder, for samples with fewer constituent tree layers, we only adopt the same number of GAT layers to encode; for samples with more constituent tree layers, we prune them to three layers.

	\section{Discussion about phrase segmentation term}\label{phrase_seg}
	We firstly provide more cases about the phrase segmentation terms in this section. For each case, the aspects are displayed in \textbf{bold} and phrase segmentation words are \underline{underlined} between the corresponding two aspects:
	
	1) However, we went for \textbf{lunch} and were the only ones eatting there \underline{and yet} the \textbf{service} seemed eager for use to be done and to get out.
	
	2) We were so excited since I was reading great review of this \textbf{place}\underline{, however} we were disappointed with the \textbf{taste} \underline{of the} \textbf{food}.
	
	3) Then the \textbf{manager} \underline{gave us} \textbf{lemon juice} \underline{instead of} \textbf{ceasar dressing} for a ceasar salad which ruined the salad. 
	
	4) The only drawback was slow \textbf{service}\underline{, but} the \textbf{food} \underline{and} \textbf{ambiance} \underline{are} so nice that your \textbf{wait} is a ) pleasant and b ) worth it.
	
	5) Compared to the \textbf{soup} of average taste\underline{,} the \textbf{rice} is better in this restaurant.
	
	The top 4 cases show that our approach can capture words, such as ``and'', ``but'', ``yet'', ``however'', ``instead of'' to help infer aspects relations. 
	
	However, we also notice there is a limitation of our method: it can only find the phrase segmentation terms within the two aspects, failing to capture some important words indicative of relations that appear in other locations. \eg in case 5), our approach capture ``,'' instead of ``compared to'', while only the latter can show the reversed sentiment of two aspects. We leave this problem as the future work, considering that our current approach is simple and can also achieve good performance. 
	
	\section{Limitations and future work}
	This section discusses some improvements that can be made in future work. 1) Our full model adopts two BERT encoders, one in \emph{Intra}-context module for encoding input text and aspects and one in \emph{Inter}-context module for encoding the phrase segmentation terms. The pros are that our \emph{Inter}-context can easily generalize to other ABSA models, taking their output aspect representations and generating the relation enhanced representations.  However, this causes the parameters of BiSyn-GAT+ up to 233M. We will consider other encoding strategies instead of simply using another BERT; 2) We notice that the label information from Con.Tree can also provide valuable information, \eg NP node and VP node, which together form the S node, may contain the aspect term and corresponding opinion words separately, as shown in Figure ~\ref{Fig:con_tree}. It is worth trying to utilize more information from Con.Tree, and we will continue to explore it in future work.

\end{document}